\documentclass[
 twocolumn,
]{ceurart}

\usepackage{listings}
\usepackage{graphicx}
\usepackage{adjustbox}
\usepackage{tabularx}
\usepackage{longtable}

\usepackage[section]{placeins}

\usepackage{tikz}
\usetikzlibrary{arrows,automata}
\usepackage{adjustbox} 
\usepackage{pgfplots}
\begin{document}

\copyrightyear{2024}
\copyrightclause{Copyright for this paper by its authors.
  Use permitted under Creative Commons License Attribution 4.0
  International (CC BY 4.0).}

\conference{Machine Learning for Cognitive and Mental Health Workshop (ML4CMH), AAAI 2024, Vancouver, BC, Canada}

\title{Toward a Reinforcement-Learning-Based System for Adjusting Medication to Minimize Speech Disfluency}


\author[]{Pavlos Constas}[%
email=pconstasm@gmail.com
]
\fnmark[1]

\author[]{Vikram Rawal}[%
email=vikram.rawal@mail.utoronto.ca
]
\fnmark[1]

\author[]{Matthew Honorio Oliveira}[%
email=matthewhonorio.oliveira@mail.utoronto.ca
]
\fnmark[1]

\author[]{Andreas Constas}[%
email=andreas.constas@mail.utoronto.ca
]
\fnmark[1]

\author[]{Aditya Khan}[%
email=aditya.khan@mail.utoronto.ca
]
\fnmark[1]

\author[]{Kaison Cheung}[%
email=siukai.cheung@mail.utoronto.ca
]
\fnmark[1]

\author[]{Najma Sultani}[%
email=najma.sultani@mail.utoronto.ca
]
\fnmark[1]

\author[]{Carrie Chen}[%
email=carrie.chen71@gmail.com
]
\fnmark[1]

\author[]{Micol Altomare}[%
email=micol.altomare@mail.utoronto.ca
]
\fnmark[1]

\author[]{Michael Akzam}[%
email=michael.akzam@mail.utoronto.ca
]
\fnmark[1]

\author[]{Jiacheng Chen}[%
email=jiachengjason.chen@mail.utoronto.ca
]
\fnmark[1]

\author[]{Vhea He}[%
email=vhea.he@mail.utoronto.ca
]
\fnmark[1]

\author[]{Lauren Altomare}[%
email=lauren.altomare@mail.utoronto.ca
]
\fnmark[1]

\author[]{Heraa Muqri}[%
email=heraa.muqri@mail.utoronto.ca
]
\fnmark[1]

\author[]{Asad Khan}[%
email=asadk.khan@mail.utoronto.ca.
]
\fnmark[1]

\author[]{Nimit Amikumar Bhanshali}[%
email=nimit.bhanshali@mail.utoronto.ca,
]
\fnmark[1]

\author[]{Youssef Rachad}[%
email=youssef.rachad@mail.utoronto.ca,
]
\fnmark[1]

\author[]{Michael Guerzhoy}[%
email=guerzhoy@cs.toronto.edu
]
\cormark[1]

\address[]{University of Toronto}

\cortext[1]{Corresponding author.}
\fntext[1]{These authors contributed equally.}

\begin{abstract}
  We propose a reinforcement learning (RL)-based system that would automatically prescribe a hypothetical patient medication that may help the patient with their mental health-related speech disfluency, and adjust the medication and the dosages in response to zero-cost frequent measurement of the fluency of the patient. We demonstrate the components of the system: a module that detects and evaluates speech disfluency on a large dataset we built, and an RL algorithm that automatically finds good combinations of medications. To support the two modules, we collect data on the effect of psychiatric medications for speech disfluency from the literature, and build a plausible patient simulation system. We demonstrate that the RL system is, under some circumstances, able to converge to a good medication regime. We collect and label a dataset of people with possible speech disfluency and demonstrate our methods using that dataset. Our work is a proof of concept: we show that there is promise in the idea of using automatic data collection to address speech disfluency.
\end{abstract}


\maketitle

\section{Introduction}

Speech disfluency is a common medical issue. It can be caused by, among other factors, conditions such as depression, anxiety, and insomnia (see Section~\ref{medlitreview}). Speech disfluency includes stuttering as well as issues like pauses that are too long, repetitions, ``false starts," and ``repairs" of previous utterances~\cite{shriberg1997prosody}.

In this paper, we propose a hypothetical Reinforcement Learning-based system for helping physicians adjust medications for people with speech disfluency. In the paper, we focus on a proof-of-concept of the system. Of course, regulatory and clinical issues would need to be addressed before actually implementing such a system. However, we show that such a system can in principle be implemented, and automatic disfluency detection can work well enough to support such a system.

On a high level, the system works as follows: the person with a speech disfluency interacts with a system that measures their speech fluency; the system tries to assign different medications in different doses to the person, to minimize their speech disfluency. Using Reinforcement Learning (RL), the system trades off exploring to find the best medication combination for the person and exploiting effective medication combinations that have been found.

We demonstrate two components of the system: a subsystem for detecting how disfluent the person's speech is, and a subsystem for minimizing the speech disfluency by finding a combination of medications that works using RL.

We train our disfluency detection system to predict the labels we assigned to clips in the dataset we collected.

To demonstrate the feasibility of the RL subsystem, we construct a patient simulation . We measure the precision with which our speech disfluency detection subsystem can measure disfluency, and obtain from the literature the plausible timespans and onset times for the effects of medications. We then run a patient simulation with plausible parameters and demonstrate that our RL subsystem can find strategies to minimize the speech disfluency of our plausibly-simulated patients.

To evaluate our subsystems, we collect a dataset of public videos of people with possible disfluencies and label the dataset using a scalable strategy that allows us to obtain precise and standardized ratings by having each video be rated by multiple raters.

The rest of the paper is organized as follows: we explain our data collection and labelling process. We then describe our disfluency rating process, and report results on that subsystem. We then describe our patient simulation process and report results of the RL system's performance on the simulated patients. For the patient simulation to be plausible, we connect the patient simulation to how precisely speech can be rated for fluency by our system and to the plausible effects of medications. (Note that if the measurement of speech fluency is too noisy and/or the medications' effect is too subtle or the onset is too long, learning would likely not be possible.) Finally, we summarize our literature search results for medications that could plausibly affect speech fluency. 

\section{Data Collection}

\subsection{Methods}
The objective of our data collection process is to obtain a series of audio samples from individuals with possible mental health-related speech disfluency, across a period of time. We collected 19 channels from searching on YouTube for mental health-related vlog channels by YouTubers, as well as the D-vlog, a dataset of channels of YouTubers with depression~\cite{yoon2022}. 

For each YouTuber represented in the videos, we scraped their channel for other videos which contained significant stretches of unedited spoken audio. Terms used to query for videos from each channel were subsets of the following keywords $\{$``depression'', ``story'', ``vlog'', ``depression vlog'', ``anxiety'', ``tested'', ``figure'', ``rambling'', ``issues'', ``anxiety vlog'', ``webcam''$\}$. For each video, only the audio was extracted. In total, we obtained $195$ audio clips. There are $9$ to $11$ audio clips for each channel, with an average of $10$ audio clips per channel. 
\subsection{Rating System}
We devised a rating system to assess the severity of the disfluency in the video data. The authors acted as raters for the videos. The 19 YouTuber channels in the dataset were examined for disfluencies in them. Raters were tasked with assessing the disfluency severity in each video on a scale of 1 to 7, which was adapted from the Stuttering Severity Instrument-Third Edition (SSI-3) \cite{riley1994}. (But note that ``disfluency" is a more general term than ``stuttering.") 

The rating process consisted of two  stages, each stage lasting approximately one week. In each stage, raters randomly received several channels and were asked to rate their audio samples from the dataset. This was arranged so that each audio sample in the dataset would have up to 3 raters. Each rater received different channels in the different stages. To ensure independent evaluation, raters were advised against sharing their assessments between each other.

Data from the initial stage was not used in our experiments. The round was used for acquainting raters with the variation of disfluency observed in the dataset. At the end of this phase, each rater was privately given summary statistics regarding their ratings in the round, including the mean and standard deviation of their ratings across audio samples, as well as a spreadsheet containing a measure of their bias for each audio sample (where bias is the distance of their rating from the mean rating across all raters for that audio sample). This process was aimed at allowing raters to recognize possible inconsistencies and biases in their ``internal model" of disfluency. The ratings from the second stage were the finalized ratings that would be used for fine-tuning our disfluency-detection system. The ratings were standardized, as described below.


\section{Rater Performance Analysis}
In this Subsection, we analyze the rater data, and show that raters are somewhat consistent in their ratings of the same clips. This indicates that we can use the standardized ratings (see below) as targets when estimating the fluency of speakers in audio clips.

\subsection{Data Model \label{datamodel}}
To assess the performance of the raters, we conducted a regression analysis. The model we utilized was $$r_{ij} = \alpha_i + \beta_j + \varepsilon_{ij}$$ where $r_{ij}$ is the rating given to audio clip $j$ by rater $i$, $\alpha_i$ is the rater bias, $\beta_j$ is the true average disfluency of the channel, and $\varepsilon_{ij}$ is the random error (see \cite{Raymond_Viswesvaran_1991} for a similar model). This model is estimated using least-squares regression. 

Using this model, the performance of the raters was assessed by randomly splitting the dataset into a 70\% training set, and a 30\% validation set. 

\subsection{Analysis}
Below, we perform an exploratory analysis of the non-standardized ratings.

We compute the Root-Mean-Square Error (RMSE) on the training set and the validation set when predicting the disfluency scores using the data model. The RMSE on the training set was $0.8/6.0$ (on a scale of 0 to 6 rather than 1 to 7 as in the input) and the RMSE on the validation set was $0.9/6.0$. The validation RMSE we would obtain if the data model predicted the average rating every time would be 1.4/6.0. The $R^2$ value of the model on the training set was 0.44,  indicating that the rater coefficients and the clip coefficients have explanatory power.

For each clip on the validation set, we compute the standard deviation of the ratings assigned by different raters to the same clip. The median standard deviation is $0.6/6.0$. This suggests that the median disagreement between raters was just over half a rating point on a given clip.  
The standard deviations are given on a scale of $6.0$ since the scores range from 1 to 7.


\subsection{Standardized Ratings}
Different raters use different standards for fluency. We therefore obtained \textit{standardized} ratings. We accomplish this by subtracting the rater bias $\alpha_i$ (see Section~\ref{datamodel}) for rater $i$ for each rating $r_{ij}$ by rater $i$. Then, when we compute the average standardized rating for every clip, we average ratings that are actually on the same scale.

\section{Disfluency Pipeline}
In this section, we describe our subsystem for assessing the disfluency of a person in the input clip.  We use Whisper~\footnote{\url{https://github.com/openai/whisper}}\cite{radford2023robust} to transcribe the audio. We then use an Auto-Correlational Neural Network-based tagger~\cite{lou2020disfluency} to tag the Whisper transcript. Finally, we fine-tune GPT-2~\cite{radford2019language} on the tagged transcript as input in order to predict the disfluency scores we assigned.

\subsection{Transcribing Audio with Whisper}
The YouTube videos are transcribed using the Automated Speech Recognition (ASR) model Whisper. Tokens such as ``uh", ``um", etc. were included in the transcript.


\subsection{Disfluency Tagging}
 The parsed text transcripts were subsequently fed into a Disfluency Tagging Auto-Correlational Neural Network (DT-ACNN) \cite{lou2020disfluency} -- a system designed to categorize each word within the text transcript as either ``fluent" or ``disfluent". In \cite{lou2020disfluency}, the Switchboard corpus of conversational speech~\cite{godfrey1992switchboard}  dataset was used. For the task of predicting a per-word ``fluent" or ``disfluent" label, the authors report a recall of 90.0\%, a precision of 82.8\%, and an F1 score of 86.2\% on the dataset. The reported results indicate the effectiveness of the DT-ACNN model in disfluency detection.

\subsection{Fine-tuning GPT-2}
 We fine-tune GPT-2 to predict the average standardized disfluency score by the rates who rated the clip from the tagged Whispter transcripts, as well as from the words-per-minute (WPM) measure.

For the regression task, we train with  embedding size 768, using the Mean Squared Error (MSE) loss, and the AdamW optimizer with parameters $\beta_1 = 0.9$, $\beta_2 = 0.999$, $\epsilon = 10^{-9}$. The token limit of GPT2 is 1024. Inputs that exceed this limit were truncated. The following hyperparameters were used during training: a learning rate of \(4.5 \times 10^{-4}\), a batch size of 4 (dictated by computational limitations), with weight decay parameter 0.01, for 50 epochs. 

P-tuning \cite{liu2023ptuning} was used.  In this approach, a soft prompt with a set of 100 tokens is introduced at the beginning of the input. These tokens aid in guiding the model during classification. The model uses a Prompt Encoder to optimize the prompt, with an encoding layer comprised of 128 units. The model performance was evaluated by
randomly splitting the dataset into an 80\% training set, and a 20\% validation set.

\subsection{Results}

\begin{figure}
\begin{flushleft}
    \begin{tikzpicture}[scale=0.79]
        \begin{axis}[
            xlabel={Epoch},
            ylabel={MSE},
            ylabel style={yshift=-15pt},
            legend entries={Training Loss, Validation Loss},
            yticklabel style={
             /pgf/number format/.cd,
             fixed,
             fixed zerofill,
             precision=2,
             /tikz/.cd
            }, 
        ]
        \addplot+ [dashed, mark=triangle*, mark options={fill=blue}, mark size=1.5] table [x=Epoch, y={Training Loss}, col sep=comma, skip coords between index={0}{2}] {gpt2_lr-0.00045_epoch50.csv};
        \addplot+ [solid, mark=square*, mark options={fill=red}, mark size=1] table [x=Epoch, y={Validation Loss}, col sep=comma, skip coords between index={0}{2}] {gpt2_lr-0.00045_epoch50.csv};
  \end{axis}
    \end{tikzpicture}
    \caption{Learning curves: disfluency prediction regression task}
    \label{fig:regression}
    \end{flushleft}
\end{figure}
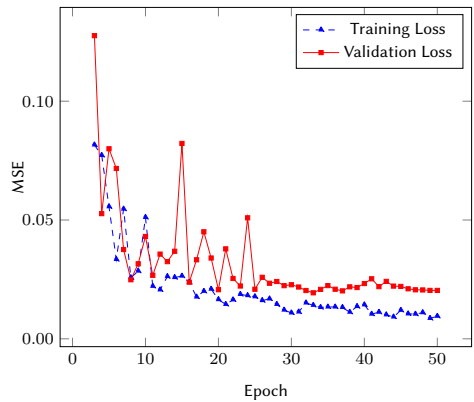

The learning curves for the disfluency prediction task are in Fig.~\ref{fig:regression}. We observe that our system is currently able to predict the validation rating to within about $0.15/6$ of the actual rating on average (the standard error is obtained by taking the square root of the MSE) for YouTubers not in the training set.

\section{Patient Simulation and Reinforcement Learning}
\subsection{Overview}
In this Section, we explore the plausibility of using RL together with signals from our speech disfluency detector to find an effective medication regimen for people with speech disfluency.

We first describe how we simulate people with speech disfluency in a plausible way. We then demonstrate that our RL algorithm could find an effective medication regimen in a plausible scenario.

\subsection{Prior Work: RL for Medication Adjustment}
Reinforcement Learning (RL) for medication adjustment has been proposed in several contexts. Oh et al. evaluate an offline RL algorithm learned on South Korea's national Health insurance system to prescribe diabetes medication \cite{oh2022reinforcement}. Javad et al. similarly propose an offline RL algorithm   \cite{javad2019reinforcement}. They measure the performance of the system based on the concordance to the prescription actually made, as well as by analyzing outcomes where the system's recommendation and the actual recommendation in the data disagreed.  Sun et al. explored anapproach for Type 2 Diabetes treatment \cite{sun2021reinforcement}. They merged a knowledge-driven model, informed by clinical guidelines, with a data-driven deep reinforcement learning model. The knowledge-driven model uses data from the Singapore Health Services Diabetes Registry which contains over 189,520 patients and their Type-2 Diabetes medication prescription to narrow down a list of viable medications to which the data-driven model applies a Deep Q Network (DQN) which also learns from the historical patient data and is used to rank the candidate medications selected by the knowledge-driven model based on expected long-term rewards.


Nemati et al. developed a clinician-in-the-loop framework for heparin dosing, leveraging data from the MIMIC-II intensive care unit database \cite{Nemati2016OptimalMD}. This study engaged an interactive agent in simulated dosing trials, learning from the outcomes to refine decision-making processes. Similarly, Anzabi Zadeh et al. utilized deep reinforcement learning in the context of warfarin dosing for patients with blood clotting issues with an emphasis on individualized dosing due to warfarin's narrow therapeutic range \cite{Anzabi_Zadeh_2023}. In this method, they frame the problem as a Markov decision process (MDP) and employ an agent within a Pharmocokinetic/Pharmacodynamic (PK/PD) model to simulate dose-responses of virtual patients in which the agent learns the best dose-duration pair through experience replay. 

\subsection{Person with Disfluency Simulation}

We model a medication administration environment, aiming to determine the most effective medication regime for people experiencing depression, anxiety, insomnia, and resulting speech fluency issues. The person's health state evolves based on Hidden Markov Models (HMMs). Each health issue (depression, anxiety, insomnia) has its unique HMM, governing how the patient's state progresses. The patient's observed speech fluency is also influenced by these health states.

The Medication object represents different types of medications, each with varying effects on the aforementioned health issues. These effects include beneficial impacts on the conditions and potential side effects. Medications have properties like dosage, half\_life, and time\_to\_effect, which dictate how they function over time.

We simulate people with disfluency by evolving the HMM state. The patient model has the following attributes:

\begin{itemize}
    \item \emph{Depression, Anxiety, Insomnia Scores}: These attributes represent the initial underlying conditions of the patient. Represented as an integer between 1 and 5.  A higher number denotes a more severe state.
    \item \emph{Depression, Anxiety, Insomnia Hidden Markov Models}: Models that represent the behaviour of how the severity of the patient’s depression, anxiety, and insomnia change over time based on their initial states and also through interaction with medicine. These directly impact observed speech fluency context.
    \item \emph{Speech Fluency Score}: Indicates the patient’s natural ability to speak fluently, modelled as a continuous value between 0 and 1.
    \item \emph{Medication Accumulation}: A list that keeps track of all medications that are currently in the patient’s system.
\end{itemize}

Alongside this, we also model an individual medication with the following attributes:

\begin{itemize}
    \item \emph{Name}: The name of the medication.
    \item \emph{Depression, Anxiety, Insomnia Effects}: Captures the medication’s average effect and variability on each condition given the standard dosage.
    \item \emph{Dosage}: The amount of medication administered relative to the standard dose. (e.g. Dosage = 1.5 means 1.5x the standard dose). This attribute scales the effects of the medication on the patient.
    \item \emph{Time to Effect}: The amount of days it takes for the medication to start showing effects.
    \item \emph{Half-Life}: The amount of days it takes for the medication dosage in the patient's system to reduce by half.
\end{itemize}

\subsection{Hidden Markov Model (HMM)}
A Hidden Markov Model (HMM) is a statistical model that represents sequences of observable data as well as hidden states. The sequences of observable data are generated based on hidden states which cannot be directly observed. Here, the ``observable data" is the patient’s speech fluency score while the ``hidden states" are the underlying depression, anxiety, and insomnia conditions that affect the severity of the disfluency. We base this model off of the fact that a patient's underlying level of depression \cite{FOSSATI200317}, insomnia \cite{JACOBS2021106106}, and anxiety \cite{WangAnxiety} have an impact on their speech fluency.

Each health condition — depression, anxiety, and insomnia — has its associated HMM. The key components of these HMMs are:

\begin{itemize}
    \item The initial probability distribution over the initial state of the condition. Initialized as a uniform distribution, indicating that any severity level is equally likely at the start.
    \item The transition matrix, which defines the probability of transitioning from one state (severity level) to another in consecutive time steps. For instance, if a patient is currently at a severity level of 3 for depression, the transition matrix, the transition matrix will dictate the probability of them improving to level 2, worsening to level 4, or remaining at level 3 in the next step.
    \item We use a Gaussian Hidden Markov Model as the observable context is assumed to be generated from a Gaussian distribution. The means and covariances define these distributions for each hidden state.
    \item Each state has a mean context emission, set to be the same int the depression, anxiety, and insomnia states.
\end{itemize}

\begin{figure}
    \centering
    \adjustbox{max width=\linewidth}{
    \begin{tikzpicture}
        \node[state] (s1) at (0,0) {1};
        \node[state] (s2) at (0,2) {2};
        \node[state] (s3) at (0,4) {3}; 
        \node[state] (s4) at (0,6) {4};
        \node[state] (s5) at (0,8) {5};

        \node[state] (e1) at (3,0) {0.8};
        \node[state] (e2) at (3,2) {0.7};
        \node[state] (e3) at (3,4) {0.6};
        \node[state] (e4) at (3,6) {0.5};
        \node[state] (e5) at (3,8) {0.4};
        
        \draw[->] (s1) to[loop left] node[midway, left] {} (s1);
        \draw[->] (s1) to[bend left] node[midway, above] {} (s2);

        \draw[->] (s1) to [bend left] node[midway, above] {output} (e1);
    
        \draw[->] (s2) to[bend left] node[midway, below] {} (s1);
        \draw[->] (s2) to[loop left] node[midway, left] {} (s2);
        \draw[->] (s2) to[bend left] node[midway, above] {} (s3);

        \draw[->] (s2) to [bend left] node[midway, above] {output} (e2);
    
        \draw[->] (s3) to[bend left] node[midway, below] {} (s1);
        \draw[->] (s3) to[bend left] node[midway, below] {} (s2);
        \draw[->] (s3) to[loop below] node[midway, below] {} (s3);
        \draw[->] (s3) to[bend left] node[midway, above] {} (s4);

        \draw[->] (s3) to [bend left] node[midway, above] {output} (e3);
    
        \draw[->] (s4) to[bend left] node[midway, below] {} (s1);
        \draw[->] (s4) to[bend right] node[midway, above] {} (s2);
        \draw[->] (s4) to[bend left] node[midway, below] {} (s3);
        \draw[->] (s4) to[loop right] node[midway, right] {} (s4);
        \draw[->] (s4) to[bend left] node[midway, above] {} (s5);

        \draw[->] (s4) to [bend left] node[midway, above] {output} (e4);
    
        \draw[->] (s5) to[bend right] node[midway, above] {} (s1);
        \draw[->] (s5) to[bend right] node[midway, above] {} (s2);
        \draw[->] (s5) to[bend right] node[midway, above] {} (s3);
        \draw[->] (s5) to[bend left] node[midway, below] {} (s4);
        \draw[->] (s5) to[loop right] node[midway, right] {} (s5);

        \draw[->] (s5) to [bend left] node[midway, above] {output} (e5);

    \end{tikzpicture}
    }
    \caption{A diagram of a Hidden Markov Model representing context emissions generated from states.}
    \label{fig:hmm}
\end{figure}
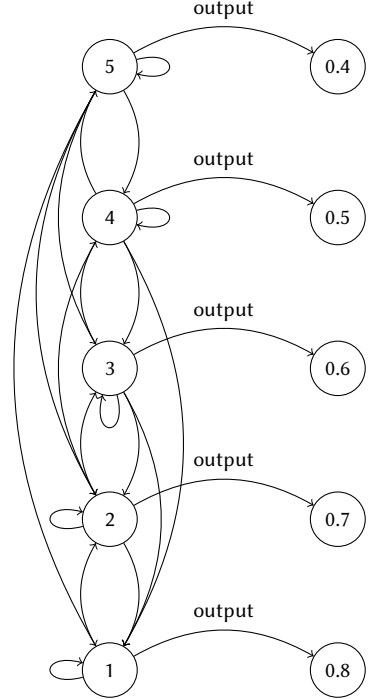

See Fig.~\ref{fig:hmm} for a diagram.






\subsection{RL Environment}

At each step, an agent can choose to administer a specific medication from the available list. The environment then evolves based on the medication's effects and the underlying psychiatric state of the patient model by the use of a transition matrix that map the current psychiatric state to a new state based on a probability distribution that models the dynamic and evolving nature of underlying psychiatric states \cite{Gauld_Depannemaecker_2023}. The agent receives a reward based on the patient's measured fluency.

We use the LinUCB~\cite{nelson2022linearizing} algorithm to learn the optimal medication strategy. The goal is to maximize the patient's speech fluency.

The effects of the medication on the patient model is implemented by applying the effects of the medication on each condition on each condition's transition matrix.

\subsection{Medication Selection Algorithm}

We use the increase in speech fluency as our reward. Speech fluency is modelled as a linear function: $$S = 0.1\cdot D + 0.2\cdot A + 0.3\cdot I + 0.4\cdot R$$where $S$ is current speech fluency, $D$, $A$, $I$ are the patient's current depression, anxiety, and insomnia scores respectively, labelled on a 5 point scale, where 1 represents no symptoms and 5 represents the most severe symptoms. $R$ is the patient's baseline fluency.

The implementation of the LinUCB algorithm observes the current state of the patient, then for each medication that is part of the environment, it estimates the reward using a linear approximation. An upper confidence bound is calculated for the estimated reward in which the medication with the highest upper confidence bound is chosen according to the equation:

$$a_t = \arg \max_{a\epsilon A} \left(x_t(\alpha)^T\theta_a + \alpha\sqrt{x_t(a)^TA_a^{-1}x_t(a)}\right)$$

where $x_t(a)$ is the feature vector for action $a$ at time $t$, $\theta_a$ is the parameter vector for action $a$ which we want to estimate, and $\alpha$ is the hyperparameter controlling the exploration/exploitation trade-off \cite{li2010contextual}. In our implementation of the LinUCB algorithm, we chose the value of $\alpha = 10.0$.

\subsection{Results}
In our simulations, we run the RL algorithm and keep track of disfluency over time.  We inject noise into the algorithm's simulated measurements of disfluency to simulate the fact that our disfluency detection system does not measure disfluency perfectly. In the experiments reported here, we inject a minimal amount of noise, corresponding to the high precision with which we can measure disfluency.

We define the success of a trial as an improvement of over $0.5\sigma$ in speech fluency. We define failure as a deterioration of over $0.5\sigma$ in speech fluency. Here, $\sigma\approx 0.1$ is the standard deviation of fluency in the dataset. 

Figures~\ref{fig:rl-results1} and \ref{fig:rl-results2} show examples of a successful and an unsuccessful run, respectively. Across 500 patient simulation runs, we found a high potential for reinforcement learning to correctly apply medication effects to reduce speech disfluency, 
with 52\% of runs showing success and 9\% of runs demonstrating failure. 

\begin{table}[htbp]
\begin{flushleft}
        \centering
        \includegraphics[scale=0.4]{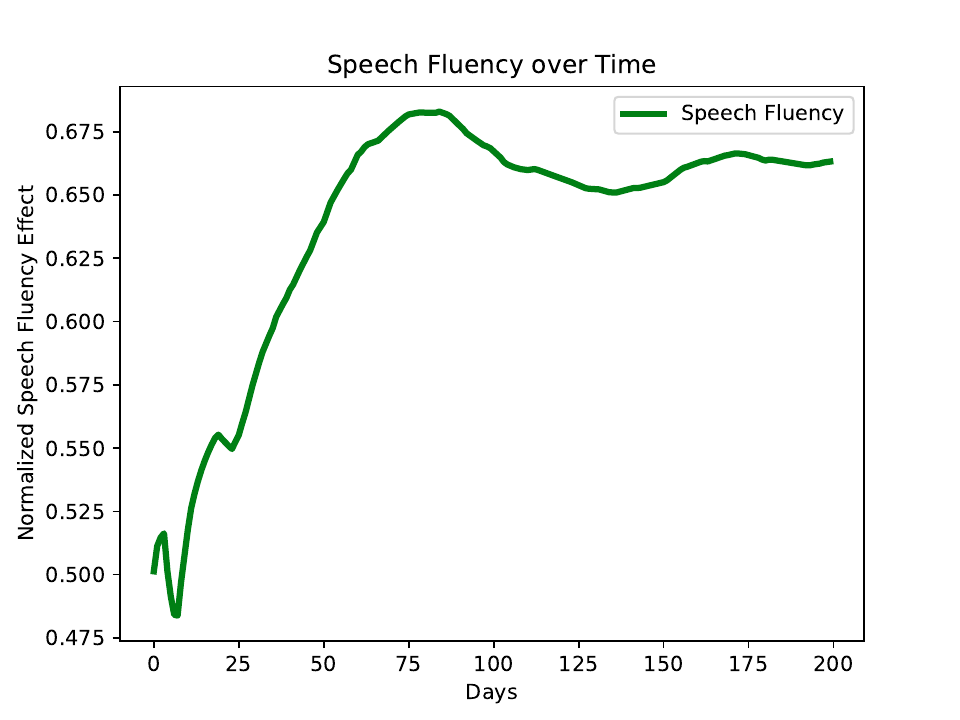}
        \captionof{figure}{Convergence to higher speech fluency.}
        \label{fig:rl-results1}

        \centering
        \includegraphics[scale=0.4]{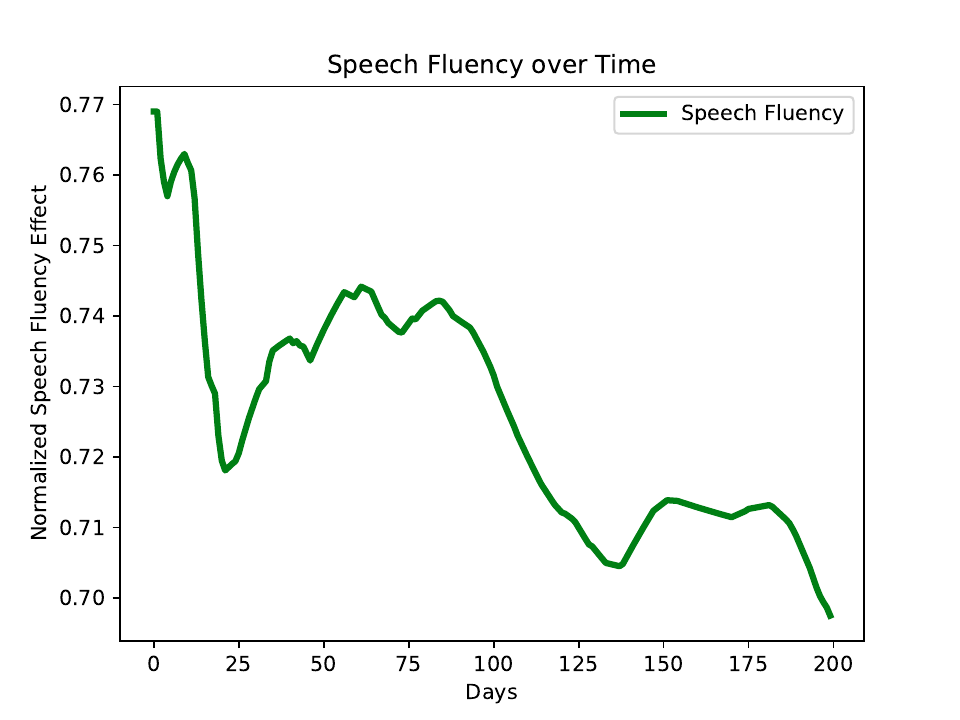}
        \captionof{figure}{Lack of convergence to higher speech fluency.}
        \label{fig:rl-results2}
    \end{flushleft}
\end{table}

The average fluency across these runs was 0.66/1.00 with a standard deviation of 0.1. The success rate of the simulations was 52\%, with a failure rate of approximately 16\%. Success and failure are defined as runs terminating greater than \(0.5\sigma\) and lower than \(-.5\sigma\) from the initial fluency level, respectively.

These results support the possibility of the use of reinforcement learning to improve speech fluency under the studied conditions. Our preliminary results indicate that if 
speech disfluency can be measured to within 10\% of the true score and the medications have plausible properties (similar to the ones seen in Section~\ref{medlitreview}), then
reinforcement learning is a possible method to dose medications that pessimize speech disfluency. However, the variability in 
outcomes and the presence of failed simulations that prompted theoretical patient deterioration indicate that further research is needed in improving the model's accuracy and understanding 
factors contributing to failures, that will be important for applying these findings in a clinical setting.

\section{Medication Literature Review \label{medlitreview}} 
A systematic literature review was conducted to determine common medications used to treat major depressive disorder (``depression") and other mental illnesses that affect speech fluency. Table \ref{tab:t1} indicates 23 medications whose onset time and response rates were used to inform the reinforcement learning simulation.

\section{Ethical Considerations}
In this paper, we outline and evaluate a proposal to adjust medications automatically in order to improve the speech fluency of a simulated patient. Although administering medications automatically is sometimes done (e.g., with Insulin pumps), this can only be done after thorough clinical trials and with informed consent from the patient. Patients must be thoroughly informed about the nature of the automated system, its potential risks and benefits, and their rights in the decision-making progress. This extends beyond initial consent, and includes ongoing consent as the system adapts medication regimens.  We believe that if safe dosage ranges can be determined, automatically adjusting medications can in principle be done. Potential upsides include better-optimized medication regimes that are found faster (or found at all). The system's reliance on sensitive patient data raises significant privacy and security concerns.  Finally, the long-term implications of integrating such a system into healthcare practices must be considered. This includes ongoing monitoring of the system’s impact on patient health outcomes, adapting to new medical insights, and updating the system accordingly.

\section{Limitations}
This paper studies the feasibility of using automated measures of disfluency together with a reinforcement learning system to adjust medications. Our model is informed by the literature, but it is still a ``toy" model. However, our results indicate that it is possible to measure disfluency to a certain precision (for some definition of ``disfluency"), and that this precision can be sufficient to serve as input to an RL system.

\section{Conclusions and Future Work}
We propose and evaluate a new idea: people with speech disfluency can have their disfluency measured automatically, and a reinforcement learning algorithm can find an optimal medication regime for them through exploration/exploitation. We have shown that the components of this system are possible to build.

In our preliminary experiments, we have shown that with an accurate enough disfluency detection system and with medications with plausible properties, a good medication regime can be found by a reinforcement learning system.

Future work includes improving our subsystems for performance, ensuring that our simulations preserve patient safety (e.g. by only trying safe dosages), and running more extensive simulations to show under what conditions RL for optimizing medications is feasible.

\bibliography{sample-ceur}

\newpage
\appendix
\onecolumn

\begin{table*}[htbp]
\newcolumntype{Y}{>{\raggedright\arraybackslash}X}
\centering
\begin{tabularx}{\linewidth}{|p{2.5cm}|Y|Y|Y|}
	\hline
	Drug Name & Medication Onset Time  & Variance in Patient Response & Treated Condition \\
	\hline\hline
	Clomipramine & 6-12 weeks \cite{wilson2019clomipramine}& Low remission rate (20\%); \cite{wilson2019clomipramine} & Obsessive compulsive disorder \cite{wilson2019clomipramine} \\
	\hline
	Imipramine & 4-5 weeks \cite{abdul2009randomized} & 80.6\% patients after 12-week treatment, compared to 48.0\% in the placebo  \cite{abdul2009randomized} &Depression, anxiety \cite{abdul2009randomized} \\
	\hline
	Selegiline & 1-2 weeks \cite{townsend2008selegiline} & 33-40\% \cite{townsend2008selegiline} &Parkinson's disease, depression \cite{moore2022selegiline} \\
	\hline
	Blonanserin & 4-6 weeks \cite{tenjin2013profile} & 52-87\% \cite{tenjin2013profile} &Schizophrenia, mania \cite{tenjin2013profile} \\
	\hline
	Venlafaxine & 4-6 weeks \cite{dean2020venlafaxine} & 67.2\% \cite{gibbons2012benefits} & Depression, anxiety, panic disorders \cite{dean2020venlafaxine} \\
	\hline
	Sertindole & 4-6 weeks \cite{cincotta2010emerging} & 85\%  \cite{cincotta2010emerging} & Schizophrenia \cite{cincotta2010emerging} \\
	\hline
	Carbamazepine & First few days \cite{maan2018carbamazepine} & 75-85\%  \cite{tolou2013quick} &Bipolar disorder, seizures \cite{maan2018carbamazepine} \\
	\hline
	Pregabalin & 3 days \cite{pande2003pregabalin} & 40\% \cite{strawn2018pharmacotherapy} &Anxiety \cite{strawn2018pharmacotherapy} \\
	\hline
	Lithium & 1-3 weeks \cite{chokhawala2022lithium} & 33\% \cite{hui2019systematic} & Bipolar disorder \cite{chokhawala2022lithium} \\
    \hline
    Ziprasidone & 1-2 weeks \cite{ionescu2016ziprasidone} & 51.85\% \cite{zhao2012efficacy} & Bipolar disorder and associated depression \cite{ionescu2016ziprasidone} \\
    \hline
    Risperidone & 4 weeks \cite{pajonk2004risperidone} & 63.4\% (4 mg), 65.8\% (8 mg) \cite{peuskens1995risperidone} & Treats irritability associated with autism and schizophrenia \cite{peuskens1995risperidone} \\
    \hline
    Gabapentin & 1-4 weeks \cite{allen2018gaba} & 17.2\% - 37.6\% (600 mg-1800mg dosage) \cite{panebianco2021gabapentin} & Epilepsy \cite{allen2018gaba} \\
    \hline
    Mirtazapine & 1 week \cite{lavergne2005onset} &67.1\% \cite{song2015efficacy} & Depression, anxiety, obsessive compulsive disorder \cite{lavergne2005onset} \\
    \hline
    Topiramate & 2-4 weeks (epilepsy), 3 months (migraines) \cite{fariba2020topiramate} & 88\% \cite{liu2020effectiveness} & Epilepsy, migraines \cite{fariba2020topiramate} \\
    \hline
    Sertraline & Within 6 weeks \cite{lewis2019clinical} & 59\% \cite{flament1999predictors} & Depression, PTSD, OCD, panic disorder, social anxiety disorder \cite{singh2019sertraline} \\
    \hline
    Citalopram & 1-2 weeks to start working, 4-6 weeks for full benefit \cite{trivedi2006evaluation} & 47\% \cite{trivedi2006evaluation} & Depression, social anxiety disorder, PTSD \cite{trivedi2006evaluation} \\
    \hline
    Duloxetine & 2-4 weeks \cite{jia2022outcomes} & 77\% \cite{jia2022outcomes} & Depression, anxiety \cite{jia2022outcomes} \\
    \hline
    Tranylcypromine & 1-2 weeks to start working, 6-8 weeks for full improvement \cite{parikh2017tranylcypromine} & 60\%-80.7\% \cite{heijnen2015efficacy} & Major depressive episodes \cite{parikh2017tranylcypromine} \\
    \hline
    Escitalopram & 1-2 weeks to start working, 6-8 weeks for full improvement \cite{waugh2003escitalopram} & 30.2\% after 2 weeks of treatment and 75.8\% after 8 weeks of treatment \cite{jiang2017efficacy} & Depression, generalized anxiety disorder (GAD), obsessive compulsive disorder (OCD) and panic attacks \cite{waugh2003escitalopram} \\
    \hline
    Quetiapine & 1-2 weeks to start working, 2-3 months for full improvement \cite{cookson2007number} & 58.2\% for quetiapine 600 mg/day and 57.6\% for quetiapine 300 mg/day after 8 weeks of treatment \cite{cookson2007number} & Schizophrenia, manic, psychotic and depressive episodes \cite{cookson2007number} \\
    \hline
    Paliperidone & 2-8 weeks for full improvement \cite{lee2011dose} & 33.6\% after 2 weeks \cite{lee2011dose} & Psychotic disorders including schizophrenia \cite{lee2011dose} \\
    \hline
    Fluoxetine & 4-5 weeks \cite{kumar2007fluoxetine} & 41\% \cite{kumar2007fluoxetine} & OCD, certain eating disorders, panic attacks \cite{kumar2007fluoxetine} \\
    \hline
\end{tabularx}
\caption{Summary of medication and variation in patient response for depression and related mental health conditions.}
\label{tab:t1}
\end{table*}

\end{document}